\pdfoutput=1

\documentclass[11pt]{article}

\usepackage[]{acl}

\usepackage{times}
\usepackage{latexsym}
\usepackage{xcolor}
\usepackage[T1]{fontenc}

\usepackage[utf8]{inputenc}

\usepackage{microtype}

\usepackage{amsmath,amsfonts,bm}
\usepackage{todonotes}
\usepackage{booktabs}
\usepackage{multirow}
\usepackage{todonotes}
\usepackage{pgfplots}
\usepgfplotslibrary{fillbetween}
\pgfplotsset{compat=1.17}

\usepackage[nointegrals]{wasysym}

\def\mF{{\bm{F}}}

\DeclareMathAlphabet{\mathsfit}{\encodingdefault}{\sfdefault}{m}{sl}
\SetMathAlphabet{\mathsfit}{bold}{\encodingdefault}{\sfdefault}{bx}{n}

\def\sZ{{\mathbb{Z}}}

\title{Effective Pre-Training Objectives for Transformer-based Autoencoders}

\newcommand*\samethanks[1][\value{footnote}]{\footnotemark[#1]}

\author{
Luca Di Liello$^{1}$\thanks{\ \ Equal contribution} , \ Matteo Gabburo$^{1}$\samethanks \ , \ Alessandro Moschitti$^{2}$ \\
$^{1}$University of Trento, $^{2}$Amazon Alexa AI \\
\texttt{\{luca.diliello,matteo.gabburo\}@unitn.it} \\ \texttt{\{amosch\}@amazon.com} \\
}

\begin{document}
\setlength{\abovedisplayskip}{1pt}
\setlength{\belowdisplayskip}{4pt}

\maketitle

\begin{abstract}
In this paper, we study trade-offs between efficiency, cost and accuracy when pre-training Transformer encoders with different pre-training objectives.
For this purpose, we analyze features of common objectives and combine them to create new effective pre-training approaches.
Specifically, we designed light token generators based on a straightforward statistical approach, which can replace ELECTRA computationally heavy generators, thus highly reducing cost. Our experiments also show that (i) there are more efficient alternatives to BERT's MLM, and (ii) it is possible to efficiently pre-train Transformer-based models using lighter generators without a significant drop in performance.
\end{abstract}


\section{Introduction}
\label{sec:intro}

Transformer-based models \citep{vaswani2017attention} require expensive hardware to be pre-trained \citep{strubell-etal-2019-energy, brown2020language}. Recently, many works focused on reducing pre-training cost \citep{lan2020albert,sanh2020distilbert,turc2019wellread}. ELECTRA, for example, proposes to train BERT as a discriminator rather than a generator \citep{clark2020electra}. They replace the Masked Language Modeling objective (MLM) \citep{devlin2019bert} with Token Detection (TD). Then, the discriminator detects if input tokens are original or fake created by a small generator network.

On the one hand, the discriminator is much more efficient than MLM. On the other hand, the use of a generator requires the pre-training of a second transformer, increasing the pre-training cost. ELECTRA has been shown to be more accurate than BERT. However, it is not clear if this superior performance is due to its innovative architecture or to the long and extensive training, which highly increases the computation cost for obtaining the final language model.

In this paper, we study pre-training strategies with respect to the trade-off between efficiency, cost, and accuracy. Theoretical efficiency and computational cost do not always align well because the latter is influenced by the underlying infrastructure and by hardware acceleration technologies (i.e., NVIDIA Tensor Cores). For this purpose, we analyze the main important components of pre-training, i.e., pre-training objectives and the algorithms with which they are applied. For example, we note that MLM needs a large classification head that spans over the whole vocabulary (which usually contains several tens of thousands of tokens), while TD requires a smaller head, which is much more efficient and uses low computation resources.

We summarize our contribution as follows:
First, we propose Random Token Substitution (RTS) and Cluster-based Random Token Substitution (C-RTS), two fast alternatives to ELECTRA's generator, which allows us to set a middle-ground in the trade-off between efficiency and accuracy.
Indeed, RTS consists in just detecting tokens that are randomly changed into others, so very low cost, while C-RTS, which is a bit more expensive than RTS, exploits the knowledge about predictions in previous iterations to select more challenging replacements.
Both our objectives increase the efficiency (20\% - 45\%) thanks to a much smaller binary classification head on top and are equally accurate to MLM on most of the tasks from a statistical significance viewpoint. We also demonstrate that, if trained for a longer time, C-RTS outperforms RTS on many benchmarks because it is a more challenging pre-training task.

Second, we propose Swapped Language Modeling (SLM), a variant of BERT's MLM that only replaces tokens with others, thus removing the special MASK token, which is responsible for BERT's pre-training/fine-tuning discrepancy~\citep{clark2020electra}.
We show that this objective increases cost with respect to RTS and C-RTS, but outperforms MLM in almost every task using precisely the same computational budget.

Finally, we empirically study the trade-offs mentioned above by pre-training standard BERT models with the proposed objectives, also comparing them with state-of-the-art architectures trained and tested on the same data. We perform an accurate comparison by evaluating our models on several natural language inference benchmarks: all tasks in the GLUE benchmark suite, ASNQ, WikiQA and TREC-QA, reporting accuracy as well as efficiency and cost. To better assess the latter, we also test the impact of objectives on smaller architectures (e.g., BERT-\textit{small}), showing that our approaches have a broader impact on those classes of models.

\begin{figure*}[ht!]
    \includegraphics[width=0.32\textwidth]{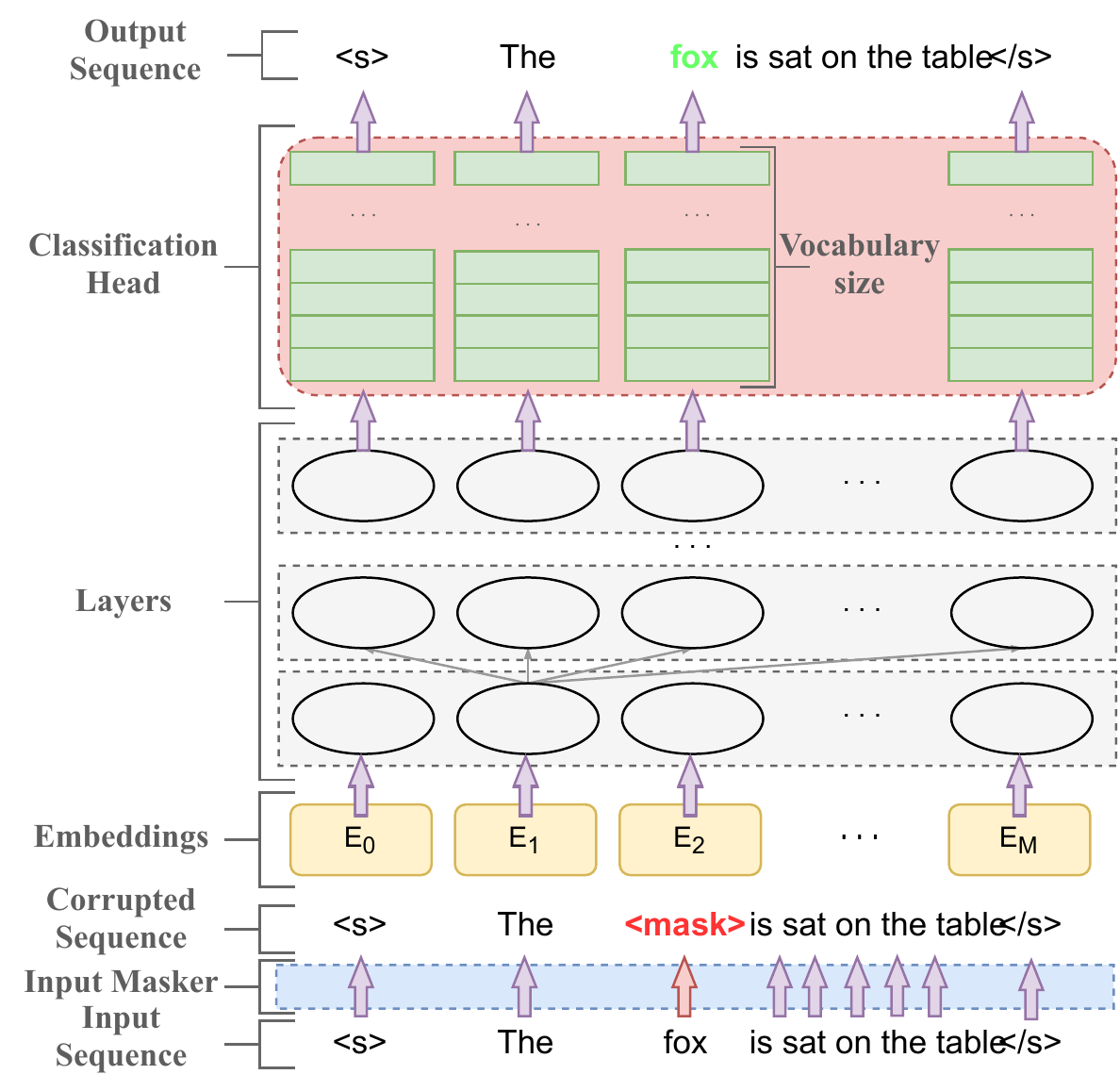}
    \hspace{\fill}
    \includegraphics[width=0.32\textwidth]{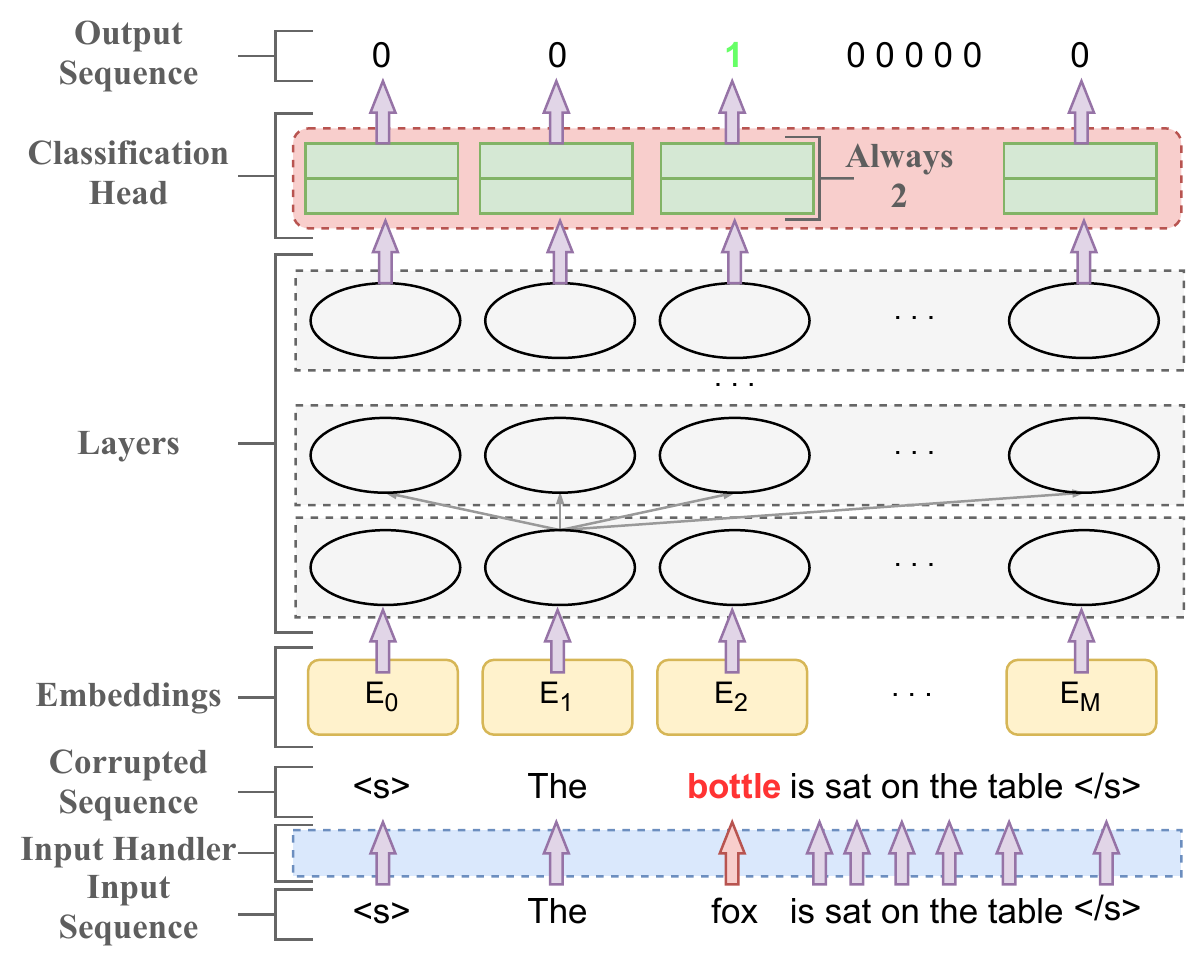}
    \hspace{\fill}
    \includegraphics[width=0.32\textwidth]{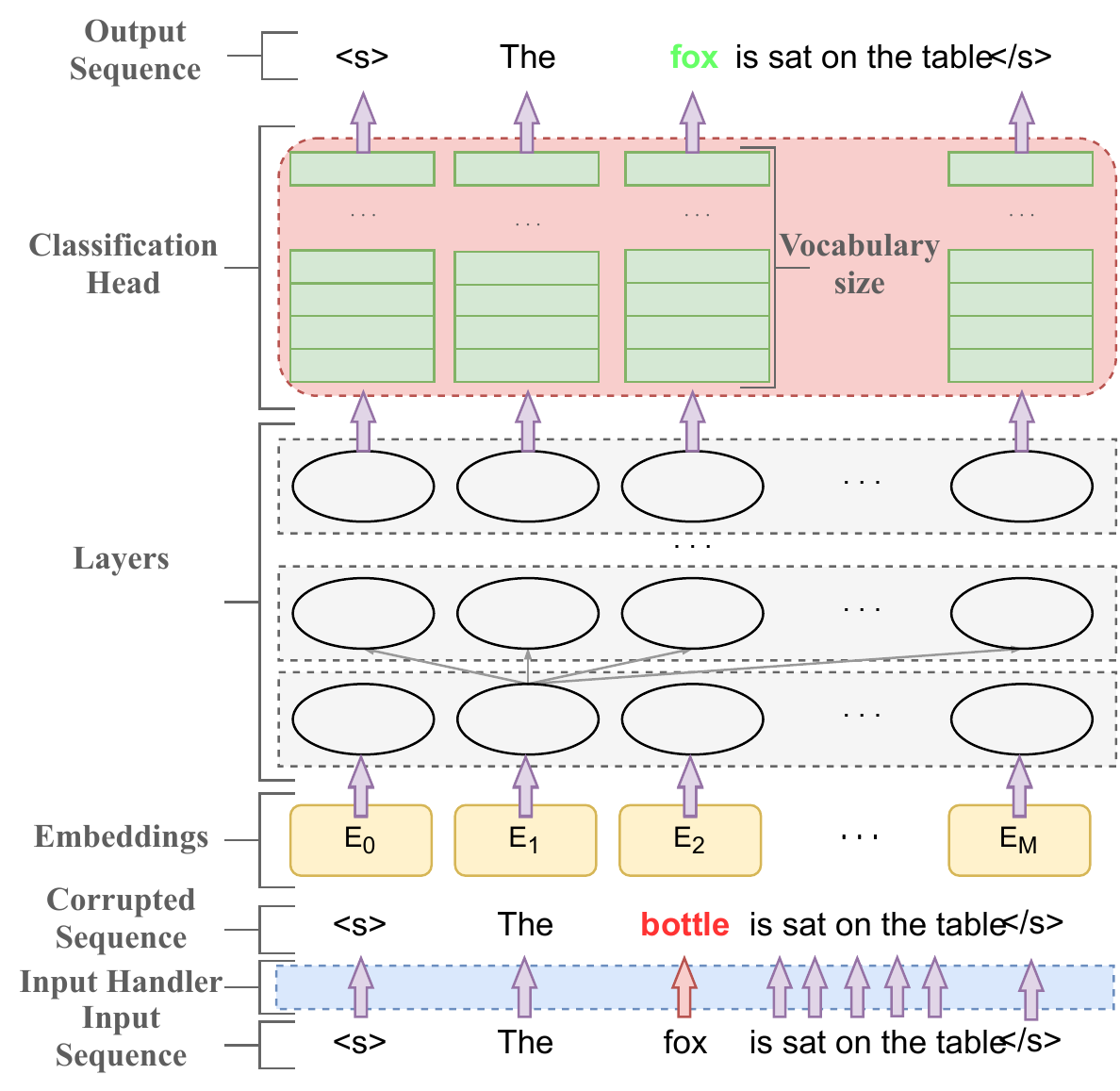}
    \caption{MLM, RTS and SLM architectures (from left to right). Notice that the classification head used by RTS is several times smaller than those used by MLM and SLM (see Appendix \ref{app:classification_heads}).}
    \label{fig:our_objectives}
    
\end{figure*}

\section{Related Work}

\label{sec:rel_work}

Many different objectives for self-supervised learning have been proposed in recent works, such as Causal Language Modeling (CLM) \citep{Radford2018ImprovingLU, radford2019language, brown2020language}, Masked Language Modeling (MLM) \citep{devlin2019bert, liu2019roberta} and Token Detection (TD) \citep{clark2020electra}, the latter used by ELECTRA, which is composed by a generator and a discriminator. While the generator is trained with MLM to find suitable candidates to replace the special MASK tokens, the discriminator should recognize the replacements in the original text instead. After pre-training, the generator is removed, and the discriminator is used as the pre-trained language model. ELECTRA introduces many innovations: (i) the exploitation of the whole output of the discriminator to compute the loss function, thus having a stronger signal for the back-propagation; (ii) the usage of a generator network to find suitable replacements and (iii) the fact that the discriminator does not see spurious tokens such as the MASK token. The latter is a main drawback of the original BERT, as it creates input discrepancies between pre-training and fine-tuning, since the CLS representation is dependent on all input tokens thanks to the self-attention mechanism.

Other research directions to reduce training time address the architecture instead of the learning objective. In ALBERT \citep{lan2020albert}, the authors tie the weights of every Transformer layer to save GPU memory, thus enabling bigger batch sizes. However, since the expressive power of their models is reduced when layers are tied, they must train for much longer. \citet{sanh2020distilbert} and \citet{turc2019wellread} instead use distillation to reduce the model size, but the pre-training is still expensive because it requires a large teacher architecture. 

Although pre-training is performed only once, it usually requires weeks and costly machines \citep{liu2019roberta, brown2020language}, so it is important to find alternative ways to pre-train transformers. \citet{tay2020efficient} provides an overview of many recent advancements in transformer efficiency.

Another successful MLM improvement regarding the objectives is SpanBERT \citep{joshi-etal-2020-spanbert}, which proposes two new objectives: Span-Masking and Span-Boundary-Objective (SBO). Specifically, the Span-Masking objective is a refined version of MLM that masks contiguous spans of text instead of single tokens, while, with SBO, they predict the span content by considering only the output representation corresponding to the tokens on the boundaries. Furthermore, in \citet{zhang-etal-2020-multi-stage}, the authors propose a technique to improve downstream performance by adapting the model to the final task while pre-training. Similarly, in \citet{diliello2022obj} they do continuous pre-training with custom objectives to better adapt the model for Answer Sentence Selection (AS2).

In T5 \citep{2020t5}, the authors propose to use deshuffling \citep{liu2019summae} to pre-train an autoregressive model. They shuffle random spans of text and ask the model to output tokens in the original order. This technique provides good results on an extensive collection of benchmarks.
However, we cannot compare with them because we focus on autoencoder architectures only.

Finally, we mention the work by \citet{izsak-etal-2021-train}, in which the authors list many optimizations that could be applied to the transformers for faster pre-training. They also claim that using larger models with the same runtime leads to better results. We focus instead on the pre-training objective efficiency and on the classification head size. Thus we state that those techniques are orthogonal to our work and could be applied along with our alternative pre-training objectives.

\section{Background on Pre-training Objectives}

\label{sec:method}

Before describing our models, we provide a detailed description of the most common token-level pre-training objectives used in the literature.

\paragraph{Masked Language Model (MLM)} was proposed by \citet{devlin2019bert}. In MLM, 15\% of the input tokens are replaced with a special mask, and the model has to predict the original value. An improvement of MLM is \textit{whole-word-masking}, in which masking is applied to every token belonging to a word and not just to independent tokens.

The model needs a large classification head for pre-training, as shown in Figure \ref{fig:our_objectives}. Its dimension is proportional to the vocabulary size, and (especially for \textit{small} models) this constitutes a significant fraction of the entire architecture parameters. In the \textit{base} architectures, the MLM head constitutes about 20\% of the model parameters while for \textit{small} models, the fraction increases to 30\%\footnote{Very often the language modelling head parameters are shared with the embedding layer}. The memory footprint of the LM head while training is about 47\% for \textit{base} and 64\% for \textit{small} models\footnote{Gradients have to be computed for every token in the vocabulary because of the final softmax layer}. See Appendix \ref{app:classification_heads} for more details. For this reason, a binary classification head (as for TD) can provide significant efficiency improvement. Moreover, sharing the parameters of the MLM classification head with the embeddings does not reduce the computational cost but leads only to marginally lower memory requirements. In the embedding layer, only a few row vectors corresponding to the actual sentence tokens are updated at every step. On the contrary, MLM's softmax continuously computes gradients for the whole output linear transformation.
\vspace{-.4em}

\paragraph{Causal Language Model (CLM)} \hspace{-.8em} is used to train autoregressive models by predicting the next token in a sequence \citep{radford2019language,Radford2018ImprovingLU}. Similarly to MLM, it requires a large classification head to output predictions over the whole vocabulary.

\vspace{-.4em}

\paragraph{Permutation Language Modeling (PLM)} \hspace{-.8em} was proposed by \citet{yang2020xlnet} to combine the generative power of autoregressive models with the bidirectional context of autoencoders. This is accomplished by permuting the input tokens and by letting the model use only the left context for the next token prediction. In this way, the model keeps the strengths of autoregressive models while exploiting the whole input sequence for better-contextualized output embedding.

\vspace{-.4em}

\paragraph{Token Detection (TD)} \hspace{-.8em} was introduced by ELECTRA~\citep{clark2020electra}, which is an architecture composed of a discriminator and a smaller generator network. First, the generator is trained with MLM and finds suitable replacements for the masked tokens, as in BERT. Then, those candidates are inserted in the original sentence, and the resulting sequence is fed to the discriminator, which classifies whether a token is original or not. TD has the advantage of computing the loss on the whole discriminator output and having a minimal memory footprint. However, the whole system is inefficient because of the presence of the generator, which is still MLM-based.

\section{Effective Pre-training Objectives}
\label{ssec:cost_efficient_pre-training_objectives}

This section presents our alternative pre-training objectives, which can potentially be applied to a wide range of Transformer-based models.

\paragraph{Random Token Substitution (RTS)}
Like ELECTRA, RTS trains a model that discriminates between original and substituted tokens. The main difference is that RTS replaces 15\% of the tokens with random alternatives, thus avoiding using computational resources to train a separate and expensive network. Besides, unlike MLM, this approach relies on a smaller classification head that is \emph{not} proportional to the vocabulary size, see Figure \ref{fig:our_objectives}.

\paragraph{Aggregated probabilities of token misclassification (C-RTS)}
The random selection applied by RTS may provide too many \emph{easy cases} to perform effective pre-training. Thus, our idea is to use a token probability distribution, inversely proportional to the classification simplicity.   More formally, let $(w_0, w_1, \dots, w_n)$ be an input sequence of tokens. A transformer model $m$ predicts $y = \{ 0, 1 \}^n$, where $y_i = 0$ indicates $w_i$ is original, and $y_i = 1$ that $w_i$ was replaced with some $w'_i$ ($w_i \rightarrow w'_i$). We aim at maximising $P( y_i = 0 \ | \ w_i \rightarrow w'_i )$, since we want to create substitutions  $w_i \rightarrow w'_i$ that are difficult to be detected by $m$.

$P( y_i = 0 \ | \ w_i \rightarrow w'_i )$ can be estimated by counting the number of failures/successes in detecting $w_i \rightarrow w'_i$ in previous iterations. While ELECTRA exploits the whole input context to create challenging replacements, our algorithm uses only the prediction history over single tokens. Storing a counter for every pair or token would generate a vast and sparse matrix given the average transformer's vocabulary size. For this reason, we partition tokens into $n$ clusters by measuring the L2 norm between the relative embedding vectors. We train the word embeddings with a \textit{word2vec} model \cite{mikolov2013w2v} on the same data used for pre-training, see Appendix \ref{app:clusering_details} for more details. After that, we use \textit{K-Means} \cite{Lloyd82leastsquares} to group the tokens into the clusters $C_1, \dots, C_n$.

We use a matrix $\mF \in \sZ^{n \times n}$, initialized with zeros, to count the difference between the failures and successes of the discriminator. While training, for each pair of tokens $(w_i \in C_a, w'_i \in C_b)$ such that $w_i \rightarrow w'_i$, we decrease $\mF_{a,b}$ by 1 if $y_i = 0$, otherwise we increment it by 1.

To maximise $P( y_i = 0 \ | \ w_i \rightarrow w'_i )$ in our approach, we select the pairs $( w_i, w'_i )$ with the highest mistake probability estimated with $\mF$. We compute the probability of selecting $w_i \in C_a$ and replacing it with $w'_i \in C_b$ as follows:
\begin{displaymath}
\resizebox{0.95\linewidth}{!}{%
$P(w_i \in C_a \rightarrow w'_i \in C_b) = P(w_i) \ P(w'_i | C_b) \ P(C_b | C_a)$%
}
\end{displaymath}
where we set (i) the probability of selecting a token in the input sequence $P(w_i)$ to 15\%; (ii) $P(w'_i | C_b)$ equal to $\frac{1}{|C_b|}$ because it is computed assuming uniform probability of choosing $w'_i$ in $C_b$; and (iii) we set $P(C_b | C_a)$ as the probability of detecting tokens from cluster $C_a$ when replaced with tokens from $C_b$. The latter is computed from $\mF$ as follows: given the candidate token $w_i \in C_a$, we define a multinomial distribution over the target clusters by indexing the $a$-th row of $\mF$. To transform the counts in $\mF_a$ into values interpretable as probabilities, we first apply the \textit{min-max} normalization and then a $\gamma$-softmax. $\gamma$ controls how the probability mass is concentrated or relaxed around the most probable cluster. For our token substitutions, we sample $C_b$ from this multinomial distribution.


We searched for the best $n$ among a reasonable set $\{30, 100, 300, 1000\}$ and the best $\gamma$ in $\{1, 2, 5, 10\}$. After preliminary experiments, we found that the best combination is $n = 100, \gamma = 2$.

\paragraph{Swapped Language Modeling (SLM)} \hspace{-.8em}
 is similar to BERT's MLM, but in this case, tokens are only randomly replaced with others and never with the special MASK. Then, unlike RTS, the model is trained to predict the original value and not discriminate between fakes and originals. SLM uses the same pre-training head as MLM; thus, it is computationally equivalent to it.

\section{Datasets}

\label{sec:dt_datasets}

This section contains the descriptions of the datasets we used for pre-training and fine-tuning.

\subsection{Pre-training datasets}

\label{{ssec:pr_datasets}}

We used Wikipedia and BookCorpus for pre-training, as in BERT \cite{devlin2019bert}. Wikipedia is a large collection of documents containing raw text for a total of about 2,500M words. We cleaned the corpus by removing lists, tables, headers, links and footers, and we considered only the passages. The BookCorpus \cite{zhu2015aligning} is composed instead of free novel books, containing approximately 800M words after cleaning. Since the original BookCorpus is not available anymore, we used the version available from the \textit{datasets} library \cite{lhoest-etal-2021-datasets}, which may result in slightly different final scores.

\subsection{Fine-tuning datasets}
\label{ssec:ft_datasets}

\paragraph{GLUE} \hspace{-.8em} (an acronym for General Language Understanding Evaluation) is a benchmark suite to test models on different NLU tasks~\cite{wang2019glue}. The collection includes datasets for Question Answering, Natural Language Inference, Question Pairs detection, Entailment Recognition and Language Acceptability. For more details about each task, see Appendix \ref{app:glue_tasks}.

\paragraph{ASNQ} \hspace{-.8em} (which stands for Answer-Sentence Natural Questions), is a dataset built for Answer Sentence Selection (AS2)~\cite{garg2019tanda}. It was derived from the Natural Questions (NQ) corpus \cite{kwiatkowski-etal-2019-natural}, which was initially designed for Machine Reading. It contains thousands of questions extracted from the Google search engine and candidate sentences retrieved from the top-ranked Wikipedia page. 
We present the details of the dataset in Table~\ref{tab:as2_datasets}.

\paragraph{WikiQA} \hspace{-.8em} is a small dataset for Answer Sentence Selection built from questions asked to the Microsoft Bing search engine~\citep{yang2015wikiqa}. Questions have been manually paired with answers taken from Wikipedia articles and labelled as relevant or not. We train on the whole training split, but we validate and test in a ``clean'' setting: questions having only positive or only negative candidates are removed.
More details in Table~\ref{tab:as2_datasets}.

\paragraph{TREC-QA} \hspace{-.8em} is another popular benchmark used for AS2~\cite{wang-etal-2007-jeopardy}. The dataset is created from the TREC-8 to TREC-13 tracks of Question Answering. As in \cite{garg2019tanda}, we train on the large \textit{train-all} split, but we do validation and testing only on the questions that have at least a positive and a negative answer. Note that \textit{train-all} contains more noise and questions that do not have positive answers. However, it is larger than \textit{train} and allows for a more stable fine-tuning. Training with \textit{train-all} and testing on clean data is the standard setting in TREC-QA. We add more details in Table~\ref{tab:as2_datasets}.

\begin{table}[t]
\small
\centering
\begin{tabular}{llcc}
\toprule
    \textbf{Dataset} & \textbf{Split} & \textbf{\# Questions} & \textbf{\# Candidates} \\
    
    \midrule
    \multirow{2}{*}{ASNQ}       & Train   & 57,242 & 20,377,568  \\
                                & Dev     & 2,672  & 930,062   \\

    \midrule
    \multirow{3}{*}{WikiQA}     & Train & 2,118 & 20,360  \\
                                & Dev     & 122   & 1,126 \\
                                & Test    & 237   & 2,341 \\
    \midrule

    \multirow{3}{*}{TREC-QA}    & Train   & 1,226 & 53,417  \\
                                & Dev     & 69    & 1,343   \\
                                & Test    & 68    & 1,442   \\

    \bottomrule

\end{tabular}
\caption{\small Statistics for the three AS2 datasets that we considered (ASNQ, WikiQA, and TrecQA). Notice that we use the "clean" setting for the dev and test splits WikiQA and TrecQA, while for ASNQ we use the original splits proposed in \cite{garg2019tanda}}
\label{tab:as2_datasets}
\end{table}


\begin{table*}[ht]
    \small
    \centering
    \resizebox{0.95\linewidth}{!}{%
    \begin{tabular}{lcccccccccc}
        \toprule

        \multirow{2}{*}{\textbf{Model}} & \textbf{CoLA} & \textbf{MNLI} & \textbf{MRPC} & \textbf{QNLI} & \textbf{QQP} & \textbf{RTE} & \textbf{SST-2} & \textbf{STS-B} & \textbf{AVG} & \multirow{2}{*}{\textbf{Time}} \\
        & matt. corr. & acc & acc & acc & acc & acc & acc & spear & \% & \\ 
        \toprule


        BERT-B / MLM+NSP $\clubsuit$ & 53.0 & 82.7 & 82.8 & 89.6 & 88.2 & 62.8 & 91.2 & 80.6 & 78.9 & $\times$ 1.00 \\
        \midrule
        BERT-B / MLM                 & 53.7 & 83.6 & 82.6 & 89.9 & 89.1 & 63.1 & 92.3 & 83.6 & 79.7 & $\times$ 1.00 \\
        BERT-B / RTS                 & 57.3 & 82.6 & 81.3 & 89.3 & 88.9 & 66.2 & 91.7 & 82.2 & 79.9 & $\times$ 0.81 \\
        BERT-B / C-RTS               & 57.3 & 82.8 & 81.9 & 89.4 & 88.6 & 60.9 & 91.0 & 82.2 & 79.3 & $\times$ 0.82 \\
        BERT-B / SLM                 & 57.0 & 83.3 & 83.1 & 89.7 & 88.9 & 65.0 & 92.3 & 83.8 & 80.4 & $\times$ 1.00 \\

        ELECTRA-B / TD               & 60.6 & 83.6 & 84.1 & 90.2 & 89.0 & 69.1 & 92.4 & 85.5 & 81.8 & $\times$ 1.05 \\

        \bottomrule
    \end{tabular}
    }
    \caption{
    \small 
    Results on the GLUE test set. Notice that the model with $\clubsuit$ is the same as Table \ref{tab:glue_dev}. We took the best models on the development set and submitted them to the GLUE leaderboard. Also, in this case, we don't do best model selection from a pool of candidates, and we do not use ensemble models. Results on the dev.~set and significance tests are available in Appendix \ref{app:glue_hparams}.
    }
    \label{tab:glue_test}
\end{table*}

\section{Experimental Setting}
\label{sec:experiments}
In these experiments, we compare the cost and accuracy of our models with the state-of-the-art methods on GLUE and several AS2 benchmarks. Finally, we summarize results derived from previous work. Then, we provide a description of the training set as well as our pre-training and fine-tuning experiments and results.

\begin{table*}[ht]
    \centering
    \resizebox{\linewidth}{!}{%
    \begin{tabular}{l|cc|cc|cc|cc|cc|c}
        \toprule

        \multirow{2}{*}{\textbf{Model}} & \multicolumn{2}{c|}{\textbf{WikiQA}} & \multicolumn{2}{c|}{\textbf{TREC-QA}} & \multicolumn{2}{c|}{\textbf{ASNQ} $\xrightarrow{}$ \textbf{WikiQA}} & \multicolumn{2}{c|}{\textbf{ASNQ} $\xrightarrow{}$ \textbf{TREC-QA}} & \multicolumn{2}{c|}{\textbf{ASNQ}} & \multirow{2}{*}{\textbf{Time}} \\
        & MAP & MRR & MAP & MRR & MAP & MRR & MAP & MRR  & MAP & MRR & \\

        \toprule
        BERT-B / MLM+NSP $\clubsuit$    & 82.8 {\tiny(0.9)} & 84.2 {\tiny(1.0)}                         & 87.2 {\tiny(0.9)} & 92.9 {\tiny(1.0)} & 87.9 {\tiny(0.4)} & 89.3 {\tiny(0.4)}                         & 89.1 {\tiny(0.4)} & 93.4 {\tiny(0.5)}                         & 66.8 {\tiny(0.1)} & 73.2 {\tiny(0.2)} &        $\times$ 1.00 \\
        \midrule
        BERT-B / MLM                    & 79.9 {\tiny(1.1)} & 81.2 {\tiny(1.2)}                         & 86.4 {\tiny(0.7)} & 91.5 {\tiny(0.7)} & 88.9 {\tiny(1.0)} & 90.2 {\tiny(1.0)}                         & 87.6 {\tiny(0.9)} & 90.6 {\tiny(1.1)}                         & 65.5 {\tiny(0.2)} & 72.2 {\tiny(0.3)} &        $\times$ 1.00 \\
        BERT-B / RTS                    & 78.5 {\tiny(2.5)} & 80.1 {\tiny(2.5)}                         & 86.6 {\tiny(1.5)} & 91.8 {\tiny(1.5)} & 87.9 {\tiny(0.5)} & 89.3 {\tiny(0.5)}                         & 88.5 {\tiny(0.5)} & \underline{93.4} {\tiny(0.4)}             & \underline{64.6} {\tiny(0.1)} & \underline{71.1} {\tiny(0.2)} &        $\times$ 0.81 \\
        BERT-B / C-RTS                  & 79.0 {\tiny(1.9)} & 80.6 {\tiny(1.6)}                         & 87.0 {\tiny(0.8)} & 91.8 {\tiny(1.1)} & \underline{87.1} {\tiny(0.6)} & \underline{88.3} {\tiny(0.6)} & \underline{88.7} {\tiny(0.4)} & \underline{92.9} {\tiny(0.7)} & \underline{64.7} {\tiny(0.1)} & \underline{71.5} {\tiny(0.1)} &        $\times$ 0.82 \\
        BERT-B / SLM                    & 80.2 {\tiny(1.5)} & 81.7 {\tiny(1.6)}                         & 87.3 {\tiny(1.3)} & 92.1 {\tiny(1.5)} & 87.7 {\tiny(0.8)} & 89.3 {\tiny(0.7)}                         & 87.9 {\tiny(0.6)} & 91.0 {\tiny(0.6)}                         & 65.8 {\tiny(0.3)} & 72.7 {\tiny(0.3)} &        $\times$ 1.00 \\
        ELECTRA-B / TD                  & \underline{81.8} {\tiny(1.6)} & \underline{83.2} {\tiny(1.6)} & 86.8 {\tiny(1.4)} & 92.2 {\tiny(1.5)} & 88.4 {\tiny(0.4)} & 89.8 {\tiny(0.4)}                         & \underline{88.9} {\tiny(0.3)} & \underline{92.0} {\tiny(0.5)} & \underline{64.9} {\tiny(0.3)} & 71.7 {\tiny(0.4)} &        $\times$ 1.05 \\
        \bottomrule
    \end{tabular}
    }
    \caption{\small Results on WikiQA and TREC-QA datasets with single-task fine-tuning and after the transfer step on ASNQ. We also report the results on the dev.~set of ASNQ optimizing MAP. The NSP loss of the original BERT mainly improves the results without the transfer step. After the transfer on ASNQ, which trains especially the CLS token representation, the difference with our MLM-based BERT is much smaller. We show the standard deviation after 5 runs with different initialization seeds in rounded brackets. We underline results that are significantly different from the BERT-B / MLM baseline after a two-sided T-Test with a significance level equal to 95\%.
    }
    \label{tab:qa_results_new_test}
\end{table*}

\subsection{Models}
We applied every objective to the same BERT \cite{devlin2019bert} architecture to make a fair comparison. Furthermore, since pre-training time is not proportional to the number of parameters but to the number of computations, we also measure the floating point operations required to do pre-training, as in \cite{clark2020electra}. FLOPS indicates the number of mathematical operations performed on the underlying hardware and are independent of the used accelerator (CPU, GPU or TPU) and the model size.

\subsection{Pre-training}
\label{sec:pre_training}

\paragraph{\textit{Base} models} We pre-trained a model for every alternative objective in the same setting as BERT-\textit{base} \cite{devlin2019bert} to perform a fair comparison. More precisely, we pre-trained models on the English Wikipedia and the BookCorpus dataset for 900K steps with a maximum sequence length of 128 tokens and another 100K steps at 512 with an uncased vocabulary. This saves much pre-training time because the computational complexity of the attention is quadratic in the sequence length. 

We use Adam and a triangular learning rate for optimization with a peak value of $10^{-4}$ and 10K warm-up steps. We use a batch size of 256 examples. More details are given in Appendix \ref{app:hparams_pre-training}.

Since ELECTRA models require more FLOPS because of the generator, we reduce the number of steps proportionally to the presence of the additional generator, as in \cite{clark2020electra}. Thus, we pre-trained ELECTRA for a total of 766K steps, of which 689K with a maximum sequence length of 128 tokens and the remaining 77K at 512.

\paragraph{\textit{Small} models}
We pre-trained 4 models using the \textit{small} architecture defined by \cite{clark2020electra} and MLM, SLM, RTS and C-RTS as objectives. We pre-train small models with the same data and hyper-parameters for the \textit{base} models but using a larger batch size of 1024 for 500K steps and always using a reduced maximum sequence length of 128 tokens.

\subsection{Fine-tuning}
\label{ssec:finetuning}

We evaluate the effectiveness of the pre-training objectives described in Section \ref{sec:method} by fine-tuning on four benchmark datasets introduced in Section \ref{ssec:ft_datasets}.


\paragraph{\textbf{GLUE}} We use the same hyper-parameters used in \cite{liu2019roberta}. More details are given in Appendix \ref{app:glue_hparams}. We measure Spearman and Matthews's correlation coefficients for STS-B and CoLA respectively, and accuracy for all the other tasks. For every model, we take the best checkpoint on the development set and evaluate it on the GLUE Leaderboard.

\paragraph{\textbf{ASNQ}} We train our models on ASNQ using a batch size of 2048 with a learning rate of $1\times10^{-5}$ for 12 epochs with early stopping on the development set. Since most question-answer pairs are shorter than 128 tokens, we use this as the maximum sequence length. We measure the performance using Mean Average Precision (MAP) and Mean Reciprocal Rank (MRR).

\paragraph{\textbf{WikiQA \& ASNQ $\rightarrow$ WikiQA}} Following the approach of TANDA \cite{garg2019tanda}, we fine-tune our models directly on WikiQA and again on WikiQA but after a transfer step on ASNQ. In particular, for each model, we run a hyperparameter search to obtain the best results. We search for the best batch-sizes in $\{32, 64, 128\}$, the best learning rates in $\{2\times10^{-6}, 5\times10^{-6}, 1\times10^{-5}, 2\times10^{-5}\}$, and we always use a maximum sequence length of 128 for up to 40 epochs. We repeat each experiment 10 times with different seeds to also report the results' standard deviation. We evaluate the model's performance on the test set using MAP and MRR. 

\paragraph{\textbf{TREC-QA \& ASNQ $\rightarrow$ TREC-QA}} For these 2 tasks, we adopted the same strategy used for WikiQA and ASNQ $\rightarrow$ WikiQA. We repeat each experiment 10 times with different seeds; in this case, we evaluate with the same metrics above. 





\begin{table*}[ht]
    \centering
    \small
    \resizebox{0.9\linewidth}{!}{%
    \begin{tabular}{l|cc|cc|cc|c|c}
        \toprule

        \multirow{2}{*}{\textbf{Model}} & \multicolumn{2}{c|}{\textbf{WikiQA}} & \multicolumn{2}{c|}{\textbf{TREC-QA}} & \multicolumn{2}{c|}{\textbf{ASNQ}} & \textbf{GLUE} & \multirow{2}{*}{\textbf{Time}} \\
        & MAP & MRR & MAP & MRR & MAP & MRR & AVG & \\
        \toprule
        BERT-S / MLM                    & 66.9 {\tiny(1.7)}   & 68.2 {\tiny(1.6)}                           & 80.9 {\tiny(2.2)} & 85.8 {\tiny(2.6)} &      58.6 {\tiny(0.3)} & 66.0 {\tiny(0.4)}     & 74.1 &    $\times$ 1.00 \\
        BERT-S / RTS                    & \underline{71.8} {\tiny(1.3)}   & \underline{73.5} {\tiny(1.5)}   & 81.4 {\tiny(0.5)} & 87.3 {\tiny(1.7)} &      \underline{59.8} {\tiny(0.1)} & \underline{66.9} {\tiny(0.2)}     & 75.4 &    $\times$ 0.54 \\
        BERT-S / C-RTS                  & \underline{69.4} {\tiny(1.3)}   & \underline{70.8} {\tiny(1.0)}   & 81.3 {\tiny(1.4)} & 86.5 {\tiny(0.7)} &      \underline{59.6} {\tiny(0.2)} & \underline{67.1} {\tiny(0.2)}     & 75.7 &    $\times$ 0.54 \\
        BERT-S / SLM                    & \underline{69.5} {\tiny(1.0)}   & \underline{70.8} {\tiny(1.0)}   & 81.9 {\tiny(1.1)} & 87.3 {\tiny(1.5)} &      \underline{59.2} {\tiny(0.2)} & 66.5 {\tiny(0.3)}     & 75.7 &    $\times$ 1.00 \\
        \bottomrule
    \end{tabular}
    }
    \caption{\small Results of RTS, C-RTS and SLM compared with MLM applied to \textit{small} architectures on AS2 benchmarks and GLUE test set. We underline the results that are significantly different from the BERT-S / MLM baseline after a two-sided T-Test with a significance level of 95\%. We show the standard deviation after 5 runs with different initialization seeds in rounded brackets.
    }
    
    \label{tab:small_qa_results_new_test}
\end{table*}

\section{Results}

\label{sec:results}
We first report the cost of our models in comparison with the state-of-the-art methods in Tables \ref{tab:FLOPS} and \ref{tab:small_times} for \textit{base} and \textit{small} models, respectively. Notice also that RTS and C-RTS use about half the GPU memory of MLM with \textit{base} and $\frac{1}{3}$ with \textit{small} models. Thus, the training time of the former's to complete all the training steps could have been even shorter by increasing the batch size. However, we keep a constant batch size to make a fair comparison. After that, we carry out a comparison in terms of accuracy on GLUE datasets as well as question-answering benchmarks. Additionally, we compare our models with state-of-the-art results published in previous works, pointing out the cost of training them.

\subsection{\textit{Base} models}

\label{ssec:modelaccuracy}

\paragraph{GLUE} Table \ref{tab:glue_test} shows that all the considered approaches, except for ELECTRA vanilla (MLM+TD), obtain comparable performance on GLUE. In particular, despite a small performance decrease compared with our MLM BERT, the RTS and C-RTS models require about 20\% less time to be pre-trained on the same machine. This suggests that using a smaller classification head greatly impacts the training time. The difference could be even broader when considering models using larger vocabularies such as RoBERTa \cite{liu2019roberta} or models with a smaller number of Transformer layers, such as DistilBERT \cite{sanh2020distilbert} or tiny-BERT \cite{turc2019wellread}.

It is also worthwhile mentioning that: (i) our BERT-SLM model achieves better performance than MLM BERT (+0.7\%), and the original BERT (+1.5\%) on average, confirming that removing the MASK token during pre-training is important; (ii) ELECTRA provides superior performance when fine-tuned on small datasets such as CoLA, while in other tasks it shows similar accuracy to SLM and MLM.

For the sake of completeness, we also show the results on the development set of the GLUE datasets in Appendix \ref{app:glue_hparams}.

\paragraph{QA benchmarks} We report the performance obtained by the models on a wide range of QA tasks in Table \ref{tab:qa_results_new_test}. The results on the dev.~set of ASNQ are obtained by fine-tuning all models in the same setting. SLM provides the highest results among all models using token-level objectives, scoring even higher than ELECTRA.

On WikiQA and TREC-QA, RTS and C-RTS have mixed results compared to MLM. In some QA tasks, RTS performs slightly better than  MLM, while, in others, slightly lower. To show that there is no statistical significant difference in performance between the two models, we do the following test: we split the test set of TREC-QA and WikiQA in 10 parts, and we test the performance of both models on all mini-batches. Finally, we take the difference in performance between the two models on every mini-batch, and we report the mean and standard deviation. We discover that MLM is better than RTS on ASNQ $\rightarrow$ WikiQA by 1.4~$(\pm 3.8)$ points while, similarly, RTS outperforms MLM on ASNQ $\rightarrow$ TREC-QA by 1.1~$(\pm 2.9)$. The high std.~dev.~makes the small difference between models insignificant. In contrast, RTS requires 20\% less time to be fully pre-trained.

C-RTS shows minor but consistent improvements over RTS in most QA tasks, especially when trained longer (see Appendix \ref{app:rtsvscrts}), also featuring a much smaller std.~dev.~when fine-tuned without the transfer step on ASNQ, which generally reduces the variability of results \cite{garg2019tanda}. Moreover, similarly to GLUE, the model trained with SLM obtains small but consistent advantages over MLM in MAP and MRR on three benchmarks out of four.

We stress the fact that the only other differences between our BERT models and the original by \citet{devlin2019bert} are the BookCorpus pre-training dataset and the additional NSP loss. We use only MLM to provide a fair comparison between token-level objectives. NSP may improve results slightly because it improves the sentence-level representation already while pre-training. This is especially true when models are not trained as long as RoBERTa or ELECTRA, which dropped NSP because it became insignificant for them, and may even hurt performance after many training steps.

\begin{table*}[ht]
    
    \centering
    \resizebox{\linewidth}{!}{%
    \begin{tabular}{lccccc}
        \toprule
        \textbf{Models} & BERT-B / MLM & BERT-B / RTS & BERT-B / C-RTS & BERT-B / SLM & ELECTRA-B / TD \\
        \midrule
        \textbf{LM head complexity} & $O(bs \times L \times |V|)$ & $O(bs \times L)$ & $O(bs \times L)$ & $O(bs \times L \times |V|)$  & $O(bs \times L \times |V|)$ \\
        \textbf{FLOPS}  & $1.61\times10^{19}$ & $1.54\times10^{19}$ & $1.54\times10^{19}$ & $1.61\times10^{19}$ & $1.98\times10^{19}$ \\
        \textbf{Training time} & 3d 14h & 2d 22h & 2d 23h & 3d 14h & 3d 18h \\
        \bottomrule
    \end{tabular}
    }
    \caption{\small FLOPS used to pre-train each \textit{base} model, language modelling head complexity and real training time on the same machine. $bs$ stands for batch size, $L$ for the input sequence length and $|V|$ is the vocabulary size, equal to about 30K tokens in BERT models. Notice that BERT-RTS and BERT-C-RTS use less memory thanks to the smaller binary classification head (also potentially allowing for larger batch sizes.)}
    \label{tab:FLOPS}
\end{table*}

\begin{table*}[ht]
    \centering
    \resizebox{0.8\linewidth}{!}{
        \begin{tabular}{lcccccc}
            \toprule
            \textbf{Models} & BERT-S / MLM & BERT-S / RTS & BERT-S / C-RTS & BERT-S / SLM \\
            \midrule
            \textbf{LM head complexity} & $O(bs \times L \times |V|)$ & $O(bs \times L)$ & $O(bs \times L)$ & $O(bs \times L \times |V|)$ \\
            \textbf{FLOPS}  & $1.83\times10^{18}$ & $1.64\times10^{18}$ & $1.64\times10^{18}$ & $1.83\times10^{18}$ \\
            \textbf{Training time} & 1d 7h & 17h & 17h & 1d 7h \\
            \bottomrule
        \end{tabular}
    }
    \caption{\small FLOPS used to pre-train each \textit{small} model, language modelling head complexity and real training time on the same hardware. See the caption of Table \ref{tab:FLOPS} for more details.}
    \label{tab:small_times}
\end{table*}

\subsection{\textit{Small} models}

Table \ref{tab:small_qa_results_new_test} shows that RTS and C-RTS on \textit{small} models outperform MLM in most tasks. In addition, RTS also improves over SLM by a wide margin in two of the three considered benchmarks (WikiQA and ASNQ). Besides, thanks to the smaller architecture, RTS greatly impacts the pre-training time. In particular, Table \ref{tab:small_times} shows that \textit{small} models exploiting RTS or C-RTS for pre-training consume half the resources used by MLM. We controlled the training of RTS and C-RTS on a small fraction of the pre-training set used for validation. We discovered that the last-epoch F1 scores of RTS and C-RTS in detecting replaced tokens were 96.3 and 94.7, respectively. This confirms that C-RTS is a more challenging task and may be well suited for longer pre-training. RTS, instead, would be more easily solved, thus providing weaker loss signals to the model.

\subsection{Models cost}

Tables \ref{tab:FLOPS} and \ref{tab:small_times} show the training compute and time required by several models. FLOPS are significant indicators but may reflect different practical performances if the underlying hardware implements special acceleration for some operations. In fact, the training times on our NVIDIA A100 GPU are not perfectly proportional to the model's FLOPS. For example, RTS and C-RTS are much faster to be pre-trained in practice in comparison to MLM and even more than the theoretical FLOPS difference thanks to the smaller memory footprint. Additionally, even by reducing the number of training steps of ELECTRA proportionally to the additional weight of the generator network, as suggested by the authors, ELECTRA still uses more computing than BERT for pre-training.

\subsection{Better modelling or just more computing?}

We aim to produce models that require less computing budget to achieve performance similar to MLM-based models. Our results show that the performance improvement is logarithmic in the size of the pre-training dataset. At the same time, there is no statistically significant difference between our computationally lighter objectives and more expensive models such as ELECTRA. Indeed, Table \ref{tab:large_models} shows that top-performing architectures outperform MLM-based models only when trained on much more data. For example, ELECTRA-\textit{Base} uses 21 times more resources than BERT-\textit{Base}, while RoBERTa uses 53 times more. It is impressive the fact that to reach a score of 90.0 on GLUE, a model has to be trained for 2000 times the original BERT.

\begin{table}[ht]
\vspace{1em}
    \centering
    \small
    \resizebox{\columnwidth}{!}{%
    \begin{tabular}{l@{\hspace{\tabcolsep}}c@{\hspace{\tabcolsep}}l}
        \toprule
        \textbf{Model} & \textbf{GLUE} & \textbf{FLOPS} \\ 
        \toprule
        BERT-S / RTS                                & 75.4  & $\times$ 0.10 \\
        BERT-S / MLM                                & 74.1  & $\times$ 0.12 \\
        BERT-S / SLM                                & 75.7  & $\times$ 0.12 \\
        BERT-B / RTS                                & 79.9  & $\times$ 0.81 \\
        BERT-B / MLM                                & 79.7  & $\times$ 1 \\
        BERT-B / SLM                                & 80.4  & $\times$ 1 \\
        BERT-L / MLM+NSP \cite{devlin2019bert}      & 83.3  & $\times$ 12 \\
        ELECTRA-B / TD \cite{clark2020electra}      & 85.7  & $\times$ 21 \\
        RoBERTa-B / MLM \cite{liu2019roberta}       & 86.3  & $\times$ 53 \\
        ELECTRA-L / TD \cite{clark2020electra}      & 88.6  & $\times$ 194 \\
        RoBERTa-L / MLM \cite{liu2019roberta}       & 88.8  & $\times$ 200 \\
        ALBERT-L / MLM+SOP \cite{lan2020albert}     & 90.0  & $\times$ 1937 \\
        \bottomrule
    \end{tabular}
    }
    \caption{Large models comparison on the GLUE test set. We report the average accuracy over the different tasks of Table \ref{tab:glue_test}, while FLOPS refers to pre-training.}
    \label{tab:large_models}
    \vspace{-1em}
\end{table}

\subsection{Does clustering really matter?}
\label{ssec:crtsvsrts}

We compared our two efficient approaches (RTS and C-RTS) to understand whether selecting more challenging replacement tokens could really improve the model performance on the downstream tasks on a long run. In order to perform this analysis, we pre-trained two \textit{small} models with both objectives using a different setting. In particular, we increased the sequence length to 512 and trained for 200K steps with a batch size of 1024, thus letting the models (i) see much more tokens and (ii) compute attention scores over long sequences. Then, we evaluated them on five different benchmarks (WikiQA, TREC-QA, ASNQ, MRPC and QNLI). We chose these datasets because they cover a wide range of tasks (AS2, Paraphrasing and NLI) and a wide range of sizes, from the 3.6K examples of MRPC to the 20M of examples in ASNQ.
The results shown in Table \ref{tab:rts_vs_rts} clearly underline that in every experiment, C-RTS achieves better performance than RTS. For example, it scores between 1 and 3\% points over RTS in MAP on the three AS2 datasets. Moreover, it provides more stable results, as it can be seen from the standard deviation across all experiments. We claim that the advantages of C-RTS over RTS derive from a more difficult pre-training objective, which is slower to converge and provides better loss signals in the last training epochs. More details are given in Appendix \ref{app:rtsvscrts}.

\begin{table}[ht]
    \centering
    \resizebox{0.8\linewidth}{!}{%
    \begin{tabular}{lccc}
        \toprule
        \textbf{Dataset} & \textbf{Metric} & \textbf{RTS} & \textbf{C-RTS} \\
        \midrule
        WikiQA  & MAP & 72.2 {\tiny(0.6)}    & 75.0 {\tiny(2.4)} \\
        TREC-QA & MAP & 84.3 {\tiny(2.4)}    & 85.4 {\tiny(1.1)} \\
        ASNQ    & MAP & 59.9 {\tiny(0.2)}    & 60.7 {\tiny(0.1)} \\
        MRPC    & Accuracy & 81.5 {\tiny(1.5)}    & 84.2 {\tiny(0.1)} \\
        QNLI    & Accuracy & 86.9 {\tiny(0.3)}    & 86.9 {\tiny(0.1)} \\
        \bottomrule
    \end{tabular}%
    }
    \caption{\small Performance comparison between RTS and C-RTS on five different benchmarks. The results reported for WikiQA and TREC-QA are on the test set, while for ASNQ, MRPC and QNLI, we report the highest score on the development set. We show the standard deviation after 5 runs with different initialization seeds in rounded brackets.
    }
    \label{tab:rts_vs_rts}
    
\end{table}

\section{Discussion and Conclusion}
\label{sec:conclusion}

In this work, we studied several alternative methods to pre-train Transformer models. Our approaches aim at designing pre-training objectives that (i) match the results of well-known methods using fewer resources or (ii) outperform previous techniques with the same computing budget. This translates into shorter training, lower memory usage, and the possibility of increasing the batch size. Among other benefits, more efficient models can reduce carbon footprint and infrastructure costs. Notice that this advantage is even higher for smaller models since the MLM classification head is not shrunk proportionally to the architecture size \cite{turc2019wellread}, thus it increases its weight on the computational complexity for smaller models. Moreover, we demonstrate that the MASK token is useless and similar or even better results can be achieved using only token substitution. We reiterate that our objectives could be easily applied to many different transformer models; we chose the BERT setting due to computational resource constraints and easy reproducibility. In Appendix \ref{app:neg_results} we provide an overview of negative results. Finally, we show that recent models' performance improvements are mostly driven by longer training phases rather than by more refined architectures.

We evaluated our approaches on several datasets, such as GLUE, WikiQA, TREC-QA and ASNQ. The results show that RTS and C-RTS match the accuracy of MLM in most tasks, requiring a lower amount of computational effort (20\% less), while SLM outperforms MLM in most tasks. C-RTS also shows a lower accuracy in detecting replaced tokens, meaning that the task is harder and better suited for longer pre-training sessions.

In addition, we tested our pre-training objectives on smaller transformer architectures. In this scenario, RTS and C-RTS obtained better performances than MLM by requiring half of its time to be pre-trained. For example, RTS outperforms MLM by almost 5 MAP points on WikiQA. This last finding is potentially helpful for training transformer models from scratch with limited resources.

In the future, we plan to explore combinations of our new techniques with efficient architectures such as ALBERT to take advantage of a lighter structure and more effective pre-training objectives.


\section{Limitations}

Pre-training of large language models is an expensive operation: powerful hardware must be reserved for many days and a lot of energy is consumed.
In this paper we explore different pre-training objectives to both save computational power and increase the performance. However, despite the improvements that we propose, the pre-training of language models still requires an incredible amount of resources. For these reasons and our computational constraints, we worked only with \textit{small} and \textit{base} architecture, leaving \textit{large} models as future work.

We trained many language models only on English training data; however, we did not explore the benefits of our alternative objectives if applied in a multilingual setting, but we believe our approaches could be easily extended to other languages with similar morphology.

Finally, we benchmark our language models on a wide range of tasks, such as Question Similarity, Answer Sentence Selection, Natural Language Inference, etc.,~but we omitted several other tasks for space limitation.


\bibliography{anthology,custom}
\bibliographystyle{acl_natbib}

\clearpage

\appendix

\section*{Appendix}

\section{Pre-training details}
\label{app:hparams_pre-training}
We pre-trained on the cleaned versions of the BookCorpus and the English Wikipedia.

For the optimization of both the \textit{small} and the \textit{base} models, we use Adam with a learning rate equal to $10^{-4}$, $\epsilon = 10^{-8}$, $\beta_1 = 0.9$ and $\beta_2 = 0.999$. The learning rate scheduler is designed to warm up for 10K steps and then decrease linearly. We use a batch size of 256 examples for the \textit{base} models and 1024 for the smalls. Finally, we apply a constant weight decay rate of 0.01, and the dropout probability is set to 0.1.

\section{Frameworks \& Infrastructure}
\label{app:infrastructure}

We implemented every model taking advantage of the HuggingFace Transformers library \cite{wolf-etal-2020-transformers}, (ii) PyTorch-Lightning for the training framework and the distributed training algorithm \citep{falcon2019pytorch} and TorchMetrics for classification and AS2 evaluation metrics \cite{Detlefsen2022}.

We performed our pre-training experiments for every model on 8 NVIDIA A100 GPUs with 40GB of memory each, using \textit{fp16} for tensor core acceleration.

\section{GLUE tasks}
\label{app:glue_tasks}

\begin{table*}
    \resizebox{\linewidth}{!}{%
    \centering
    \begin{tabular}{l|cccccccc|c}
        \toprule

        \multirow{2}{*}{\textbf{Model}} & \textbf{CoLA} & \textbf{MNLI} & \textbf{MRPC} & \textbf{QNLI} & \textbf{QQP} & \textbf{RTE} & \textbf{SST-2} & \textbf{STS-B} & \textbf{AVG} \\
        & matt. corr. & acc & acc & acc & acc & acc & acc & spear & \% \\ 
        \toprule


        
        BERT-S/ MLM     & 42.2 {\tiny(0.9)}              & 78.8 {\tiny(0.2)}              & 80.8 {\tiny(0.9)}             & 85.7 {\tiny(0.6)}             & 88.6  {\tiny(0.1)}             & 58.6 {\tiny(1.3)}             & 89.4 {\tiny(0.2)}             & 84.4 {\tiny(0.2)} & 76.1 \\
        BERT-S / RTS    & \underline{51.2} {\tiny(1.8)}  & \underline{79.9} {\tiny(0.2)}  & 81.5 {\tiny(0.7)}             & \underline{87.9} {\tiny(0.1)} & \underline{89.4} {\tiny(0.1)}  & \underline{61.2} {\tiny(0.6)} & \underline{88.4} {\tiny(0.4)} & \underline{85.2} {\tiny(0.2)} & 78.1 \\
        BERT-S / C-RTS  & \underline{51.6} {\tiny(0.9)}  & \underline{79.8} {\tiny(0.1)}  & 81.3 {\tiny(1.0)}             & \underline{87.0} {\tiny(0.4)} & \underline{89.4}  {\tiny(0.1)} & \underline{61.6} {\tiny(1.4)} & 89.5 {\tiny(0.2)}             & \underline{85.3} {\tiny(0.3)} & 78.2 \\
        BERT-S / SLM    & \underline{45.8} {\tiny(0.5)}  & 79.1 {\tiny(0.2)}              & \underline{83.3} {\tiny(0.4)} & 86.3 {\tiny(0.3)}             & \underline{89.0}  {\tiny(0.1)} & \underline{60.8} {\tiny(1.4)} & \underline{88.6} {\tiny(0.4)} & \underline{86.0} {\tiny(0.2)} & 77.4 \\

        \midrule
        BERT-B / MLM+NSP $\clubsuit$ & 57.6 {\tiny(1.8)}              & 84.3 {\tiny(0.4)} & 82.3 {\tiny(1.3)} & 91.0 {\tiny(0.7)} & 91.0  {\tiny(0.2)} & 68.9 {\tiny(1.4)} & 92.6  {\tiny(0.1)} & 89.1 {\tiny(0.3)} & 82.1 \\
        \midrule
        BERT-B / MLM                 & 58.1 {\tiny(1.0)}              & 83.4 {\tiny(0.2)}             & 87.5 {\tiny(0.5)}             & 90.2 {\tiny(0.3)}             & 90.9  {\tiny(0.1)}            & 67.4 {\tiny(1.2)}             & 92.2 {\tiny(0.3)} & 87.8 {\tiny(0.3)}             & 82.2 \\
        BERT-B / RTS                 & 58.1 {\tiny(1.1)}              & \underline{82.7} {\tiny(0.2)} & 87.6 {\tiny(1.0) }            & \underline{89.4} {\tiny(0.3)} & 90.9  {\tiny(0.1)}            & 68.5 {\tiny(1.4)}             & 91.5 {\tiny(0.3)} & \underline{86.6} {\tiny(0.4)} & 81.9 \\
        BERT-B / C-RTS               & 57.4 {\tiny(0.7)}              & \underline{82.0} {\tiny(0.3)} & \underline{84.2} {\tiny(0.4)} & 89.6 {\tiny(0.2)}             & \underline{90.6} {\tiny(0.1)} & 66.6 {\tiny(2.4)}             & 91.5 {\tiny(0.2)} & 87.0 {\tiny(0.2)}             & 81.1 \\
        BERT-B / SLM                 & \underline{59.6} {\tiny(1.0)}  & 83.4 {\tiny(0.2)}             & 87.5 {\tiny(0.4)}             & 89.9 {\tiny(0.2)}             & 91.0  {\tiny(0.1)}            & \underline{69.2} {\tiny(1.2)} & 92.1 {\tiny(0.1)} & 87.6 {\tiny(0.3)}             & 82.5 \\

        \midrule
        ELECTRA-B / TD               & 63.4 {\tiny(1.3)} & 83.7 {\tiny(0.2)} & 87.2 {\tiny(0.8)} & 90.4 {\tiny(0.1)} & 91.2 {\tiny(0.1)} & 74.6 {\tiny(1.4)} & 91.4 {\tiny(0.4)} & 88.5 {\tiny(0.2)} & 83.8 \\
        \bottomrule
    \end{tabular}
    }
    \caption{
    \small 
    Results on GLUE dev.~set for both \textit{base} and \textit{small} models (we use the suffixes -B and -S). The symbol $\clubsuit$ indicates the official BERT-\textit{base} uncased pre-trained model released by \cite{devlin2019bert}, which uses an additional NSP loss during the pre-training. We trained every other model in the same setting. For each task, we fine-tune 5 times and take the best model on the development set. We do single task fine-tuning without best model selection or using ensemble models like in \cite{clark2020electra}.
    For each group of our BERT models, we underline results that are statistically different from the MLM baseline model by doing a statistical T-Test with a significance level equal to 95\%.
    }
    \vspace{-0.8em}
    \label{tab:glue_dev}
\end{table*}

The collection includes: (i) two datasets to test performance in paraphrasing capabilities, one composed of questions (QQP) pairs and the other of the sentence pairs (MRPC); (ii) a dataset for question-answer entailment (QNLI) derived from the SQUAD dataset \cite{rajpurkar2016squad}; (iii) three datasets for textual entailment (RTE, MNLI and WNLI); (iv) a single dataset (STS-B) to test the model on textual similarity; (v) a dataset (SST-2) to evaluate performance on sentiment analysis and finally (vi) a dataset to check linguistic acceptability (CoLA). As in \cite{devlin2019bert} and \cite{clark2020electra}, for stability reasons we avoid testing on WNLI.

We report the results obtained in the development set in Table \ref{tab:glue_dev} for both the \textit{base} and the \textit{small} models. 

\section{GLUE hyperparameters and development set results}
\label{app:glue_hparams}
We used a batch size of 16 with $lr=1\times10^{-5}$ for CoLA and MRPC, a batch size of 16 with a learning rate of $2\times10^{-5}$ for RTE and STS-B, and a batch size of 32 and a learning rate of $1\times10^{-5}$ for MNLI, QNLI, QQP and SST-2. We set the maximum sequence length to 128 for every task. All the GLUE experiments use a triangular learning rate scheduler with 10\% of warmup. We train models using half-precision, and optimize them on the development set. Table \ref{tab:glue_dev} provides the results on the development set of GLUE for all models.

\section{Non impactful objectives}
\label{app:neg_results}

\subsection{C-RTS sampling within the same cluster}
We experimented with a simplification of C-RTS where tokens are always replaced with tokens within the same cluster. We found that it is important to maintain the possibility to sample also from other clusters because the model was able to learn how tokens are clustered after an enough large number of steps.

\subsection{Position-based techniques}
We experimented with changing the position of tokens, i.e., we masked some positional encoding in an MLM-like approach. The objective was to retrieve the position of the masked tokens in the original sentence. The classification head of this approach is smaller than MLM (slightly larger than RTS, because of the 512 possible positions in BERT), but the results were worse by 3-4\% points on GLUE.

We also defined another objective consisting of (i) shuffling input positions of some tokens, and then (ii) predicting their original position. We selected 15\% of the tokens, and we permuted them. Although the results were below MLM by 1-2\% on the GLUE average, the approach was as much fast as RTS. A combination of position-based objectives with the token-level ones is a possible future direction.

\subsection{LM head-on ELECTRA's discriminator}
Capitalizing on the good performance reached from SLM, we implemented and evaluated a version of SLM for ELECTRA. Since SLM cannot be directly applied to the ELECTRA discriminator (the predictions are only performed on the output positions corresponding to tampered tokens nullifying the task of detecting fake tokens), we propose an alternative objective called \textit{SLM-all}. In this case, the discriminator has to predict the whole input sentence, estimating which tokens were changed and predicting their original values. At the same time, it should only reproduce the input in output for unchanged inputs. As for the other objectives, we evaluated this approach on GLUE, WikiQA, ASNQ and TREC-QA but we obtained generally worst results than some of the other approaches, also with a more expensive training (1.42 times the time required by MLM with \textit{base} models). This gap in the efficiency between SLM-all and the other models is so large because the latter predicts MLM-like token for every input token. Notice that even by reducing the number of training steps of the ELECTRA model by about 25\% (to balance the presence of the additional generator with size 1/3), it uses slightly more FLOPS than BERT-MLM.
Specifically, it scores 80.02 on the GLUE average score, 78.7, 86.7, 86.7 of MAP in WikiQA, TREC-QA and ASNQ.

\section{Clustering of tokens embeddings}
\label{app:clusering_details}
First, token embeddings have been obtained by training a word2vec \cite{mikolov2013w2v} model over the same data used in pre-training. We also tried to use embeddings from an already pre-trained BERT model, but we saw no significant difference in TD accuracy. Therefore, we decided to use word2vec to provide a complete pre-training pipeline and not rely on an already pre-trained checkpoint. We used a context of 2 words on either side of the target tokens and an embedding size of 300. The training algorithm written in PyTorch \cite{NEURIPS2019_9015} took less than 10 minutes on the same GPU used for pre-training. Thus this process is not significant concerning the whole pre-training time.
We perform clustering of tokens using the K-means algorithm \cite{Lloyd82leastsquares}. We used the CPU implementation of K-means provided by the \textit{scikit-learn} library \cite{scikit-learn}, doing 20 random starts. The clustering took approximately 12 minutes on the Intel Xeon Platinum 8275CL in our machine.



\section{Classification heads}
\label{app:classification_heads}

\vspace{-0.5em}

BERT's MLM has to make predictions over the whole vocabulary. For this reason, the last layer of the model should output a probability distribution over the whole vocabulary, which usually contains about 30K tokens. In particular, BERT uses a linear layer of size $H \times |V|$ ($H$ is the hidden size of the model) followed by a softmax to generate values that could be interpreted as probabilities. On the contrary, RTS or C-RTS needs just a binary classification head to predict whether a token is original or fake. This results in a simple linear layer with size $H \times 2$. For this reason, MLM's classification head is usually tens of thousands times larger than RTS's.

\section{C-RTS vs RTS with longer pre-training.}
\label{app:rtsvscrts}
We demonstrate the superiority of C-RTS over RTS by doing a longer pre-training on a \textit{small} model. We increased the maximum sequence length from 128 to 512 and kept the same batch size of 1024. We train until both models converge and no longer improve, exploiting the same pre-training data used for the other experiments. After every epoch, we evaluate the checkpoints on various tasks: WikiQA, TREC-QA and MRPC. We avoid very large datasets such as ASNQ because, on \textit{small} models, they underline more the architectural differences than the pre-training technique.

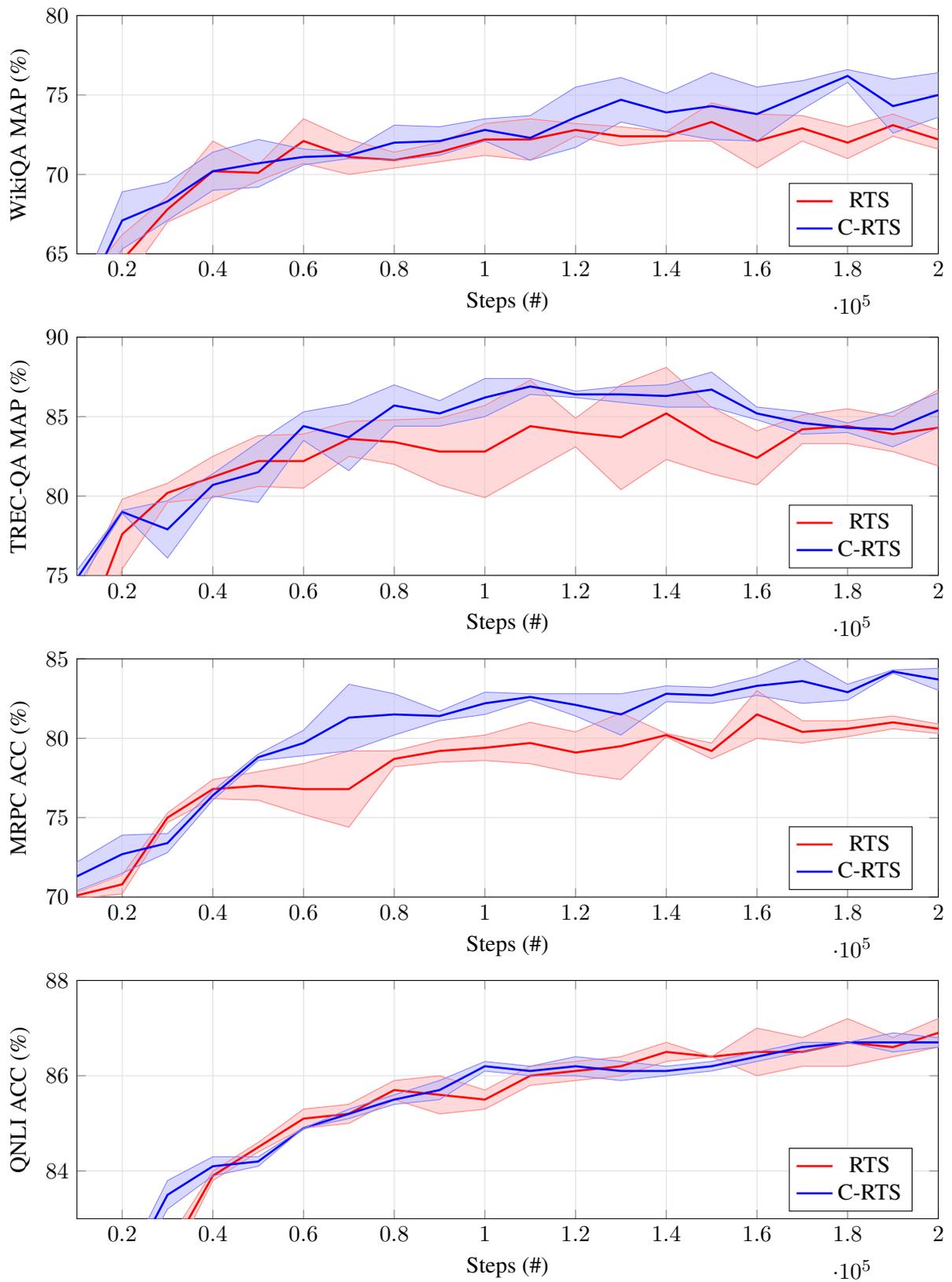
\begin{figure*}[ht]
\centering

\begin{tikzpicture}
    \begin{axis}[%
    	xlabel=Steps (\#),
    	ylabel=WikiQA MAP (\%),
    	grid=both,
    	xmin=10000, xmax=200000, xstep=10000,
        ymin=65, ymax=80,
    	minor grid style={gray!25},
    	major grid style={gray!25},
    	width=1.0\linewidth,
    	height=0.35\linewidth,
    	legend pos=south east]

    \addplot[line width=1pt,color=red] %
    	table[x=steps,y=map_avg,col sep=comma]{csvs/wikiqa_rts_topdown.csv};
        \addlegendentry{RTS};
    \addplot[name path=rts_top,color=red!50] %
    	table[x=steps,y=map_std_top,col sep=comma]{csvs/wikiqa_rts_topdown.csv};
    \addplot[name path=rts_down,color=red!50] %
    	table[x=steps,y=map_std_down,col sep=comma]{csvs/wikiqa_rts_topdown.csv};
    \addplot[red!30,fill opacity=0.5] fill between[of=rts_top and rts_down];

    \addplot[line width=1pt,color=blue] %
    	table[x=steps,y=map_avg,col sep=comma]{csvs/wikiqa_crts_topdown.csv};
        \addlegendentry{C-RTS};
    \addplot[name path=us_top,color=blue!50] %
    	table[x=steps,y=map_std_top,col sep=comma]{csvs/wikiqa_crts_topdown.csv};
    \addplot[name path=us_down,color=blue!50] %
    	table[x=steps,y=map_std_down,col sep=comma]{csvs/wikiqa_crts_topdown.csv};
    \addplot[blue!30,fill opacity=0.5] fill between[of=us_top and us_down];

    \legend{RTS,,,,C-RTS,,,,};
   \end{axis};
\end{tikzpicture}

\begin{tikzpicture}
    \begin{axis}[%
    	xlabel=Steps (\#),
    	ylabel=TREC-QA MAP (\%),
    	grid=both,
    	xmin=10000, xmax=200000, xstep=10000,
        ymin=75, ymax=90,
    	minor grid style={gray!25},
    	major grid style={gray!25},
    	width=1.0\linewidth,
    	height=0.35\linewidth,
    	legend pos=south east]

    \addplot[line width=1pt,color=red] %
    	table[x=steps,y=map_avg,col sep=comma]{csvs/trecqa_rts_topdown.csv};
    \addplot[name path=rts_top,color=red!50] %
    	table[x=steps,y=map_std_top,col sep=comma]{csvs/trecqa_rts_topdown.csv};
    \addplot[name path=rts_down,color=red!50] %
    	table[x=steps,y=map_std_down,col sep=comma]{csvs/trecqa_rts_topdown.csv};
    \addplot[red!30,fill opacity=0.5] fill between[of=rts_top and rts_down];

    \addplot[line width=1pt,color=blue] %
    	table[x=steps,y=map_avg,col sep=comma]{csvs/trecqa_crts_topdown.csv};
    \addplot[name path=crts_top,color=blue!50] %
    	table[x=steps,y=map_std_top,col sep=comma]{csvs/trecqa_crts_topdown.csv};
    \addplot[name path=crts_down,color=blue!50] %
    	table[x=steps,y=map_std_down,col sep=comma]{csvs/trecqa_crts_topdown.csv};
    \addplot[blue!30,fill opacity=0.5] fill between[of=crts_top and crts_down];

    \legend{RTS,,,,C-RTS,,,,};
   \end{axis};
\end{tikzpicture}

\begin{tikzpicture}
    \begin{axis}[%
    	xlabel=Steps (\#),
    	ylabel=MRPC ACC (\%),
    	grid=both,
    	xmin=10000, xmax=200000, xstep=10000,
        ymin=70, ymax=85,
    	minor grid style={gray!25},
    	major grid style={gray!25},
    	width=1.0\linewidth,
    	height=0.35\linewidth,
    	legend pos=south east]

    \addplot[line width=1pt,color=red] %
    	table[x=steps,y=acc_avg,col sep=comma]{csvs/mrpc_rts.csv};
    \addplot[name path=rts_top,color=red!50] %
    	table[x=steps,y=acc_std_top,col sep=comma]{csvs/mrpc_rts.csv};
    \addplot[name path=rts_down,color=red!50] %
    	table[x=steps,y=acc_std_down,col sep=comma]{csvs/mrpc_rts.csv};
    \addplot[red!30,fill opacity=0.5] fill between[of=rts_top and rts_down];

    \addplot[line width=1pt,color=blue] %
    	table[x=steps,y=acc_avg,col sep=comma]{csvs/mrpc_crts.csv};
    \addplot[name path=crts_top,color=blue!50] %
    	table[x=steps,y=acc_std_top,col sep=comma]{csvs/mrpc_crts.csv};
    \addplot[name path=crts_down,color=blue!50] %
    	table[x=steps,y=acc_std_down,col sep=comma]{csvs/mrpc_crts.csv};
    \addplot[blue!30,fill opacity=0.5] fill between[of=crts_top and crts_down];

    \legend{RTS,,,,C-RTS,,,,};
   \end{axis};
\end{tikzpicture}

\begin{tikzpicture}
    \begin{axis}[%
    	xlabel=Steps (\#),
    	ylabel=QNLI ACC (\%),
    	grid=both,
    	xmin=10000, xmax=200000, xstep=10000,
        ymin=83, ymax=88,
    	minor grid style={gray!25},
    	major grid style={gray!25},
    	width=1.0\linewidth,
    	height=0.35\linewidth,
    	legend pos=south east]

    \addplot[line width=1pt,color=red] %
    	table[x=steps,y=acc_avg,col sep=comma]{csvs/qnli_rts.csv};
    \addplot[name path=rts_top,color=red!50] %
    	table[x=steps,y=acc_std_top,col sep=comma]{csvs/qnli_rts.csv};
    \addplot[name path=rts_down,color=red!50] %
    	table[x=steps,y=acc_std_down,col sep=comma]{csvs/qnli_rts.csv};
    \addplot[red!30,fill opacity=0.5] fill between[of=rts_top and rts_down];

    \addplot[line width=1pt,color=blue] %
    	table[x=steps,y=acc_avg,col sep=comma]{csvs/qnli_crts.csv};
    \addplot[name path=crts_top,color=blue!50] %
    	table[x=steps,y=acc_std_top,col sep=comma]{csvs/qnli_crts.csv};
    \addplot[name path=crts_down,color=blue!50] %
    	table[x=steps,y=acc_std_down,col sep=comma]{csvs/qnli_crts.csv};
    \addplot[blue!30,fill opacity=0.5] fill between[of=crts_top and crts_down];

    \legend{RTS,,,,C-RTS,,,,};
   \end{axis};
\end{tikzpicture}

\caption{Performance comparison of RTS and C-RTS on WikiQA, TREC-QA, MRPC and QNLI.}
\label{fig:rtsvscrts}

\end{figure*}

\end{document}